\documentclass{article}

\PassOptionsToPackage{numbers}{natbib}

\usepackage[dandb]{neurips_2025}

\usepackage[utf8]{inputenc} % allow utf-8 input
\usepackage[T1]{fontenc}    % use 8-bit T1 fonts
\usepackage{hyperref}       % hyperlinks
\usepackage{url}            % simple URL typesetting
\usepackage{booktabs}       % professional-quality tables
\usepackage{amsfonts}       % blackboard math symbols
\usepackage{nicefrac}       % compact symbols for 1/2, etc.
\usepackage{microtype}      % microtypography
\usepackage{xcolor}         % colors
\usepackage{authblk}        % for \affil command
\usepackage{enumitem}
\usepackage{array} % Required for centering columns
\usepackage{amsmath}
\usepackage{tabularx}
\usepackage{algorithm}
\usepackage{algpseudocode}
\usepackage{titletoc}
\usepackage{amssymb}
\usepackage{multirow}
\usepackage{tcolorbox}
\usepackage{parskip}
\usepackage{xcolor}

\usepackage{pifont}
\usepackage{cleveref}
\usepackage{booktabs} 
\usepackage{array} % Required for centering columns
\usepackage{amsmath,bm}
\usepackage{adjustbox}  
\usepackage{fontawesome}
\definecolor{my_green}{RGB}{51,102,0}
\definecolor{my_red}{RGB}{204, 0, 0}
\definecolor{my_half}{RGB}{226,115,0} 
\newcommand{\halfcheck}{\textcolor{my_half}{\ding{51}\rotatebox[origin=c]{-9.2}{\kern-0.7em\ding{55}}}}
\definecolor{paired-light-blue}{RGB}{198, 219, 239}
\definecolor{paired-dark-blue}{RGB}{49, 130, 188}
\definecolor{paired-light-orange}{RGB}{251, 208, 162}
\definecolor{paired-dark-orange}{RGB}{230, 85, 12}
\definecolor{paired-light-green}{RGB}{199, 233, 193}
\definecolor{paired-dark-green}{RGB}{49, 163, 83}
\definecolor{paired-light-purple}{RGB}{218, 218, 235}
\definecolor{paired-dark-purple}{RGB}{117, 107, 176}
\definecolor{paired-light-gray}{RGB}{217, 217, 217}
\definecolor{paired-dark-gray}{RGB}{99, 99, 99}
\definecolor{paired-light-pink}{RGB}{222, 158, 214}
\definecolor{paired-dark-pink}{RGB}{123, 65, 115}
\definecolor{paired-light-red}{RGB}{231, 150, 156}
\definecolor{paired-dark-red}{RGB}{131, 60, 56}
\definecolor{paired-light-yellow}{RGB}{231, 204, 149}
\definecolor{paired-dark-yellow}{RGB}{141, 109, 49}  
\definecolor{myblue}{RGB}{218,232,252}
\definecolor{mygray}{RGB}{220,220,220}
\definecolor{mypink}{RGB}{251,49,153}

\usepackage{caption}
\usepackage{makecell}
\usepackage{colortbl}
\usepackage{enumitem}
\usepackage{caption}
\usepackage{graphicx}
\usepackage{float} 
\usepackage{wrapfig}
\usepackage{subcaption}
\usepackage{hyperref}
\usepackage{xcolor}

\usepackage{graphicx}

\newcommand{\HFIcon}{\includegraphics[height=1em]{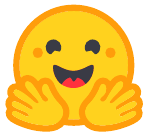}}

\title{
  \centering
  \raisebox{0.4ex}{\includegraphics[scale=0.13, bb=130 170 200 0]{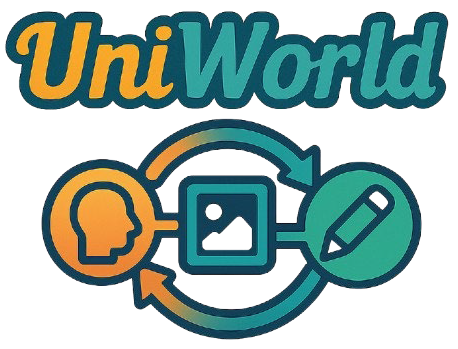}}
  \quad \quad 
  UniWorld-V1: High-Resolution Semantic Encoders for \\
  \quad \quad \quad Unified Visual Understanding and Generation
}

\author{%
  \textbf{Bin Lin}\textsuperscript{1,3},
  \textbf{Zongjian Li}\textsuperscript{1,3},
  \textbf{Xinhua Cheng}\textsuperscript{1,3},
  \textbf{Yuwei Niu}\textsuperscript{1,3},
  \textbf{Yang Ye}\textsuperscript{1,3},
  \textbf{Xianyi He}\textsuperscript{1,3},
  \textbf{Shenghai Yuan}\textsuperscript{1,3},
  \textbf{Wangbo Yu}\textsuperscript{1,3},
  \textbf{Shaodong Wang}\textsuperscript{1,3},
  \textbf{Yunyang Ge}\textsuperscript{1,3},
  \textbf{Yatian Pang}\textsuperscript{1},
  \textbf{Li Yuan}\textsuperscript{1,2,$^\dagger$}
}

\affil{
  \par
  \textsuperscript{1} Peking University, Shenzhen Graduate School, 
  \textsuperscript{2} Peng Cheng Laboratory, 
  \textsuperscript{3} Rabbitpre AI
  \\
  \par
}

\begin{document}

\maketitle
% \footnotetext{Corresponding Authors †, Equal Contributors *}

\begin{table}[ht]
  \centering
  \footnotesize
  \begin{tabular}{@{}l l@{}}
    \faGithub\quad \ Code:   & \url{https://github.com/PKU-YuanGroup/UniWorld-V1} \\[4pt]
    \HFIcon\quad Models:   & \url{https://huggingface.co/LanguageBind/UniWorld-V1} \\[4pt]
    \HFIcon\quad Data:     & \url{https://huggingface.co/datasets/LanguageBind/UniWorld-V1} \\[4pt]
  \end{tabular}
\end{table}

\begin{figure*}[h]
    \centering
    \includegraphics[width=1\linewidth]{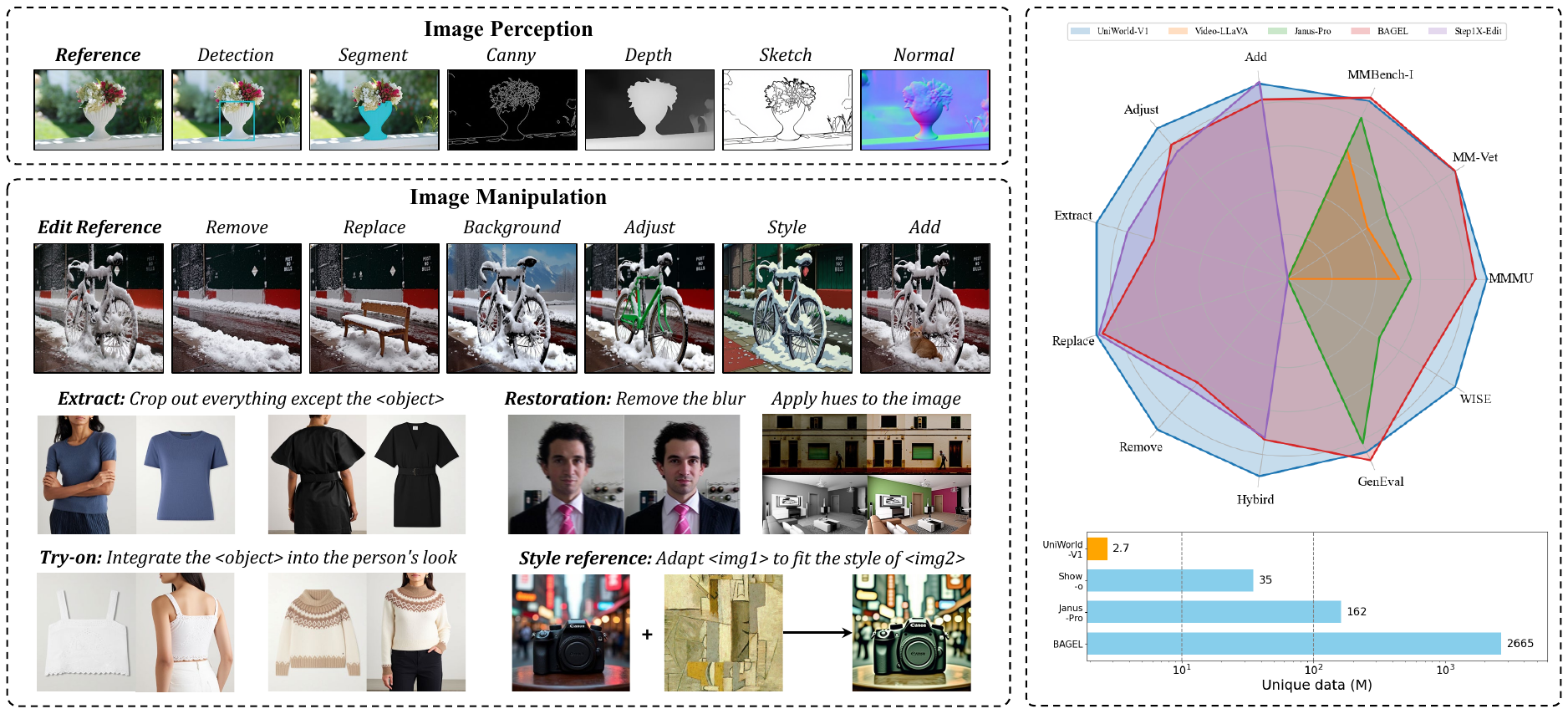}
    \caption{\textbf{Showcase of UniWorld-V1’s versatile capabilities.} The left two panels illustrate image perception and manipulation examples, and the right panel presents comparisons with state-of-the-art models and training data resources.
    }
    \label{fig:first_fig}
\end{figure*}

\begin{abstract}
% Although existing unified models deliver strong performance on vision-language understanding and text-to-image generation, their models are limited in exploring image perception and manipulation tasks, which are urgently desired by users for wide applications. 
% Recently, OpenAI released their powerful GPT-4o-Image model for comprehensive image perception and manipulation, achieving expressive capability and attracting community interests.  
% By observing the performance of GPT-4o-Image in our carefully constructed experiments, \textit{we infer that GPT-4o-Image leverages features extracted by semantic encoders instead of VAE}, while VAEs are considered essential components in many image manipulation models.
% Motivated by such inspiring observations, we present a unified generative framework named \textbf{UniWorld-V1} based on semantic features provided by powerful visual-language models and contrastive semantic encoders.
% As a result, we build a strong unified model using only 2.7M data, which demonstrates exceptional performance in image understanding, generation, manipulation, and perception tasks.
% We fully open-source our models, including model weights, training \& evaluation scripts, and datasets.
Although existing unified models achieve strong performance in vision-language understanding and text-to-image generation, they remain limited in addressing image perception and manipulation---capabilities increasingly demanded in practical applications. Recently, OpenAI introduced the powerful GPT-4o-Image model, which showcases advanced capabilities in comprehensive image perception and manipulation, sparking widespread interest. Through carefully designed experiments, we observe that \textit{GPT-4o-Image likely relies on semantic encoders rather than VAEs} for feature extraction, despite VAEs being commonly regarded as crucial for image manipulation tasks. Inspired by this insight, we propose \textbf{UniWorld-V1}, a unified generative framework built upon semantic features extracted from powerful multimodal large language models and contrastive semantic encoders. Using only 2.7M training data, UniWorld-V1 achieves impressive performance across diverse tasks, including image understanding, generation, manipulation, and perception. We fully open-source the UniWorld-V1 framework, including model weights, training and evaluation scripts, and datasets to promote reproducibility and further research.

\end{abstract}

\section{Introduction}\label{sec: introduction}

Unifying image understanding and generation has demonstrated remarkable capabilities on multi-modal models.
Recent studies \cite{tong2024metamorph,zhang2025unified,Janus} demonstrate that carefully designed architectures can be jointly optimized to perform well on both understanding and generation benchmarks. 
The remarkable visual understanding and generation capabilities exhibited by the recent GPT-4o-Image model have further energized the open-source community, leading to a surge of new unified models \cite{BLIP3-o,BAGEL} aiming to replicate its performance.
GPT-4o-Image achieves superior generation performance on various image-to-image tasks across different domains, which can be divided into two categories, including image perception tasks~\cite{sam2,yolo} (\textit{e.g.}, detection) and image manipulation~\cite{imgedit} (\textit{e.g.}, editing) tasks.

Nevertheless, most unified models are limited to image-to-language understanding tasks and language-to-image generation tasks, with very few addressing image-to-image perception and manipulation tasks.
However, attempting to tackle image perception and manipulation tasks in one model is difficult because such an all-in-one model requires multiple superior capabilities, including \textbf{(1)} the textual and visual unified understanding ability for correctly addressing user intention, \textbf{(2)} the pixel-level information maintaining ability for image reconstruction and region editing, and \textbf{(3)} the semantic extraction ability for cross-domain perception and visual conception composition.

Therefore, the requirements of various capabilities impose specific designs for the unified generation model, especially for visual feature injection.
Recent attempts including Step1X-Edit~\cite{step1x_edit} and FLUX-Kontext~\cite{FLUX-Redux} introduce variational autoencoders (VAEs) for extracting visual features and perform well on individual image editing tasks.
However, their methods encounter challenges when extended to multiple perception and manipulation tasks simultaneously due to the visual features encoded by VAEs, which imply a wealth of low-frequency information, thereby limiting their performance when facing semantic-level tasks.

Inspired by the success of GPT-4o-Image, we investigate the integration of visual features into unified generative models for image manipulation tasks, a direction that remains insufficiently explored in current research.
Therefore, we carefully construct experiments on GPT-4o-Image to infer the visual feature extraction manner that GPT-4o-Image likely adopts, and we infer that \textbf{GPT-4o-Image employs visual features extracted by semantic encoders rather than VAEs by observing from the experimental results.}

Based on our essential observation shown in Section~\ref{sec: observation}, we propose a unified generative model named \textbf{UniWorld-V1} for both image perception and manipulation tasks, which consists of pre-trained multi-modal large models for providing auto-regressive understanding tokens and pre-trained high-resolution contrastive semantic encoders for extracting visual features with both pixel-level local information and semantic-level global conceptions.

As a result, UniWorld-V1, trained on only 2.7M samples, consistently outperforms BAGEL~\cite{BAGEL} (trained on 2665M samples) on the ImgEdit-Bench~\cite{imgedit} for image manipulation. It also surpasses the specialized image editing model Step1X-Edit across multiple dimensions, including add, adjust, and extract on ImgEdit-Bench. Additionally, for text-to-image generation, UniWorld-V1 outperforms BAGEL on WISE~\cite{niu2025wise} and achieves performance comparable to GPT-4o-Image on GenEval~\cite{ghosh2023geneval}. Furthermore, beyond its strong capabilities in the aforementioned tasks, UniWorld-V1 also excels in Image Perception tasks such as detection, segmentation, and depth prediction. In summary, we are the first open-source model to achieve such comprehensive and powerful capabilities across the multimodal domain.

We summarize our primary contributions as follows:
\begin{itemize}
\item Provide insights into unified architecture design by hypothesizing, through extensive observation of GPT-4o-Image, that it likely does not adopt a VAE-based structure.
\item Propose UniWorld-V1, a unified architecture that leverages a high-resolution semantic encoder to deliver reference-image control signals. UniWorld-V1 achieves performance comparable to BAGEL using only 2.7M training samples.
\item Collect, curate, and open-source high-quality training data. We fully release the code, model weights, datasets, and training \& evaluation scripts to support and advance future research.
\end{itemize}

\begin{figure*}[!t]
    \centering
    \includegraphics[width=1\linewidth]{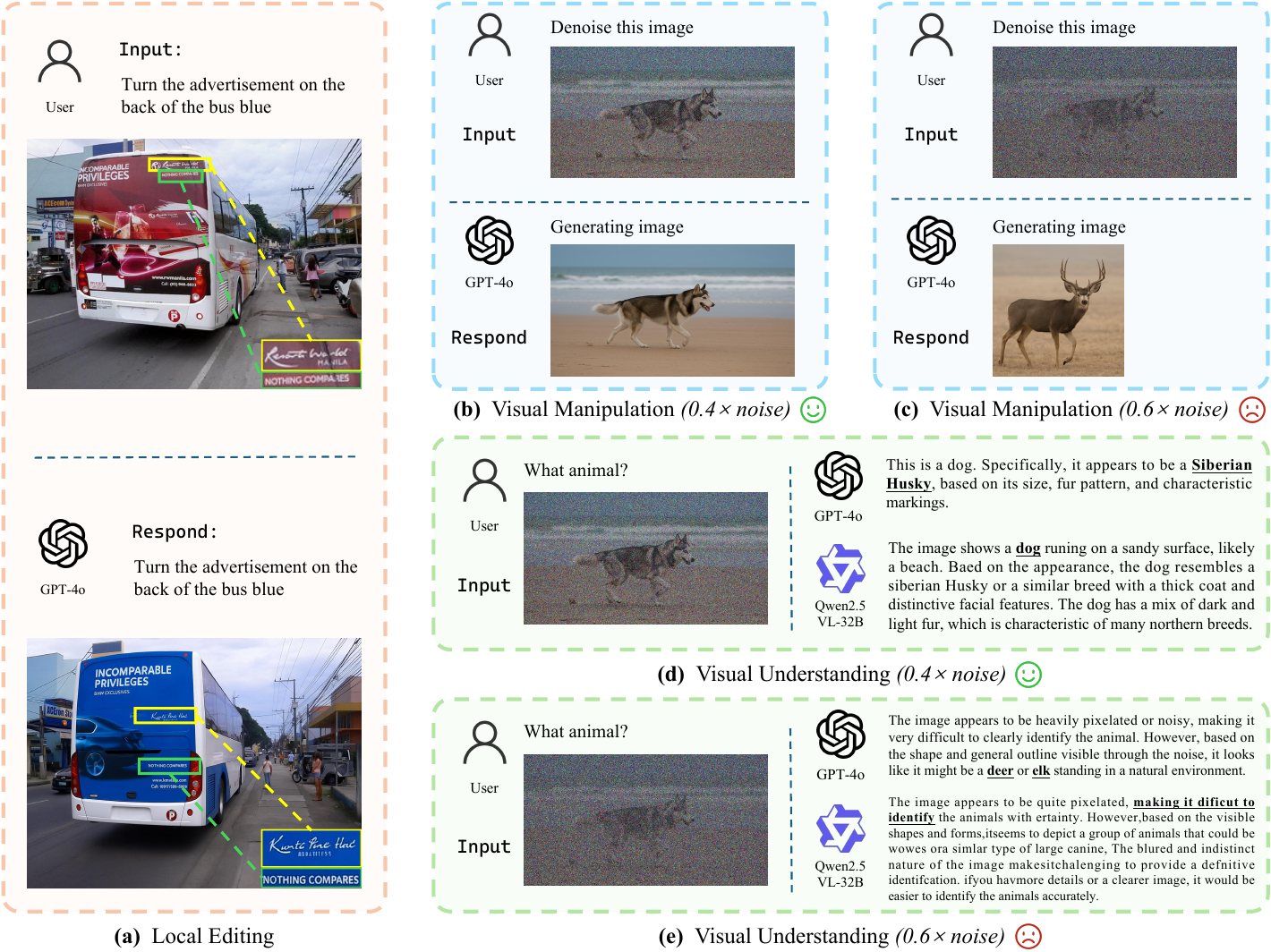}
    \caption{\textbf{Empirical Observations of GPT-4o-Image.} We verify local consistency of edits in (a). We explore the relationship between model comprehension and generation in (b)–(e), conducting observations within the GPT-4o architecture and across architectures using Qwen2.5VL-32B.
    }
    \label{figure_observation}
\end{figure*}

\section{Observation}\label{sec: observation}
GPT-4o-Image \cite{gpt4} achieves impressive performance in various image tasks in a generative way, and we divide supported image tasks into two categories: \textbf{Image Perception} (detection, segmentation, depth prediction, \textit{etc.}) and \textbf{Image Manipulation} (editing~\cite{step1x_edit,zhang2025context,chen2025multimodal}, reference style transfer~\cite{OmniConsistency}, subject consistency generation~\cite{consisid, opens2v}, \textit{etc.}).
It is generally believed within the community that GPT-4o-Image demonstrates the necessity of unifying understanding and generation by integrating an auto-regressive understanding module~\cite{zhu2023languagebind,liao2025langbridge,lin2024moe} with a diffusion-based generation module~\cite{gpt_imgeval} for addressing various image tasks with complex requirements.
However, we consider that additional visual features beyond auto-regressive tokens from understanding models should be injected to maintain image information, since we fail to train an effective generation model from solely VLM outputs.
Recent notable unified image manipulation approaches, exemplified by Step1X-Edit~\cite{step1x_edit} and FLUX-Kontext~\cite{FLUX-Redux}, simultaneously introduce variational autoencoders (VAEs) to extract image featuresa as reference image control.
Although their models perform well on individual editing tasks, our experiments show that they fail to converge when extended to multiple image perception and manipulation tasks, which indicates that the manner of additional visual feature injection remains underexplored.
To explore the visual feature injection method that GPT-4o-Image utilizes, we construct two groups of experiments and obtain several key observations (see Figure \ref{figure_observation}), from which we infer that \textbf{GPT-4o-Image likely employs visual features extracted by semantic encoders rather than VAEs.}

\noindent\textbf{Editing Experiment.}
We first require GPT-4o-Image to execute a local image editing task with the instruction: ``Turn the advertisement on the back of the bus blue'', as shown in Figure~\ref{figure_observation}~(a). 
Before the editing, both yellow-labeled and green-labeled texts are in the top-right corner of the bus. 
However, the yellow-labeled text is on the right, and the green-labeled text is on the bottom right after editing.
We claim that if GPT-4o-Image leverages VAE features that strongly preserve the low-frequency information for visual injection, the positions of texts would remain virtually unchanged, while GPT-4o-Image renders both texts into inconsistent positions.

\noindent\textbf{Denoising Experiment.}
We then corrupt a dog image with noise levels of $0.4\times$ and $0.6\times$ for GPT-4o-Image to execute an image denoising task with the instruction: ``Denoise this image'', as shown in Figure~\ref{figure_observation}~(b) \& (c).
We observe that GPT-4o-Image performs normally when the noise is small, but wrongly denoises the image as a deer when the noise level is $0.6\times$. 
We claim that VAE features preserve the low-frequency components of the input (\textit{e.g.}, global structures and contours) and lead to correct denoise results.
Besides, to figure out why GPT-4o-Image denoises the dog image as a deer, we query two multi-modal understanding models, including GPT-4o and Qwen2.5-VL, as shown in Fig.~\ref{figure_observation}~(d) \& (e). Interestingly, both understanding models caption the $0.6\times$-noised dog image as a deer, which demonstrates GPT-4o-Image is based on the prior of powerful multi-modal understanding models.

In summary, our experiments demonstrate that GPT-4o-Image more likely employs visual features extracted by semantic encoders rather than VAEs since the low-level information of the source image is not preserved after the manipulation.
Inspired by such observations, we design the architecture of our unified image perception and manipulation methods named UniWorld-V1.
\section{Method}\label{sec: method}

\begin{figure*}[h]
    \centering
    \includegraphics[width=1\linewidth]{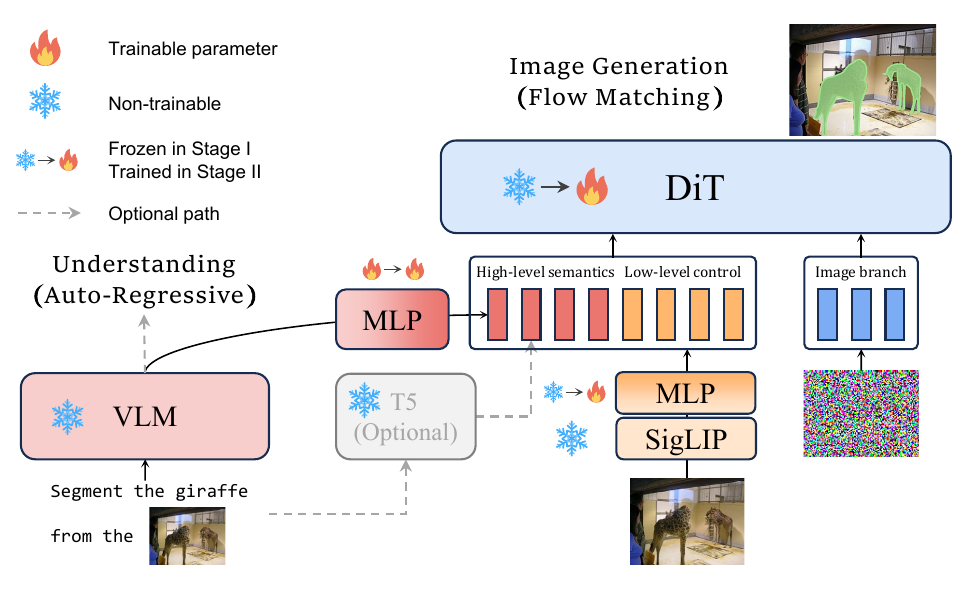}
    \caption{\textbf{Model architecture.} The model consists of a VLM, SigLIP, DiT \cite{peebles2023scalable}, and MLP connector. High-level semantics and historical state are provided by the VLM, while low-level image features are controlled by SigLIP. The understanding part uses a frozen VLM with an autoregressive approach, while the generation part is trained with flow matching. T5 (originally used for conditional injection) is optional during training or generation.
    }
    \label{fig:method}
\end{figure*}

\subsection{Model Architecture}

Based on our empirical observations, we replace the VAE-based low-level control signal with the SigLIP encoder~\cite{zhai2023sigmoid,tschannen2025siglip} (Figure~\ref{fig:method}), a contrastive vision-language model that demonstrates superior performance. We use the largest resolution variant, which is SigLIP2-so400m/14 with a fixed resolution of 512. For visual understanding, we follow prior work~\cite{MetaQuery,BLIP3-o} and use the pretrained Qwen2.5-VL-7B~\cite{Qwen25VL} as the base module. The reference image is processed by both Qwen2.5-VL-7B and SigLIP, and their outputs are concatenated as the input to the text branch of FLUX~\cite{FLUX}.

\subsection{Training Recipe}

During training, the T5~\cite{raffel2020exploring} features (the original condition used in FLUX) are optional. However, we observe that incorporating T5 features in the early stages often leads to convergence to poor local minima. Therefore, we do \textbf{NOT} recommend using T5 features early in training.

\textbf{Stage 1: Pretraining for Semantic Alignment} Due to a feature gap between VLM representations and the FLUX text branch, stage 1 focuses on aligning VLM features to T5 features. During this stage, only the MLP mapping from VLM to FLUX is trainable, while all other parameters remain frozen. Moreover, since stage 1 is solely dedicated to aligning VLM semantic features, SigLIP features are excluded. After stage 1 pretraining, the model can perform text-to-image generation and produce images that differ from the reference based on editing instructions.

\textbf{Stage 2: Fine-Tuning for Consistent Generation} We load the weights of the VLM to FLUX MLP trained in stage 1 and the MLP weights from FLUX-Redux~\cite{FLUX-Redux}, which align SigLIP features to the text branch. We unfreeze all learnable parameters in the FLUX image branch while keeping all text branch parameters frozen. Although stage 1 aligns VLM to FLUX, early in stage 2, the model still takes a shortcut by directly reconstructing the reference image. After 5,000 to 10,000 training steps, the model begins learning how to use SigLIP features as reference cues to generate images according to instructions.

\subsection{ZeRO-3 EMA} 

\begin{wrapfigure}{r}{0.4\textwidth}  
    \centering
    \includegraphics[width=\linewidth]{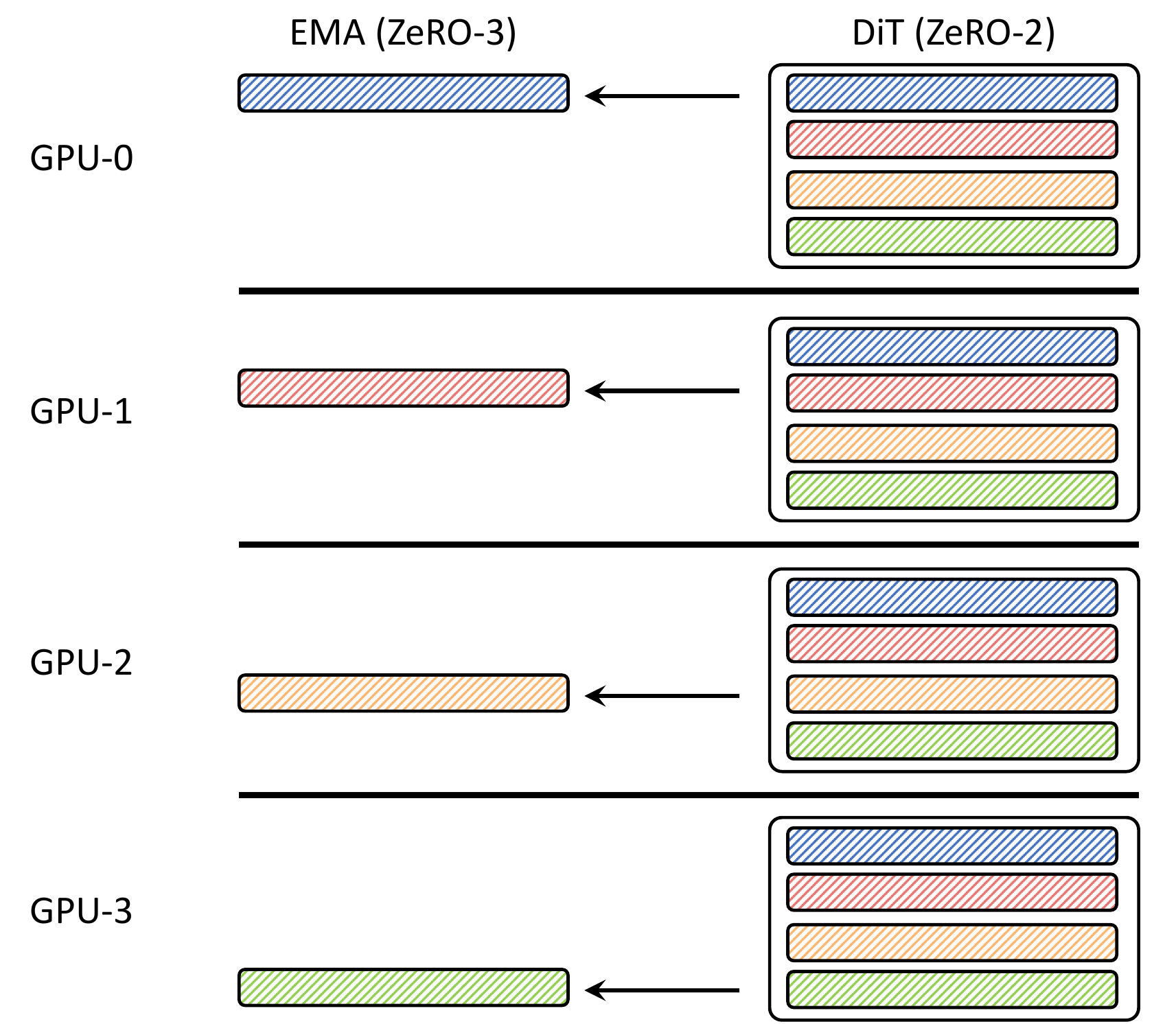}
    \caption{\textbf{Zero-3 EMA.} EMA model is initialized with Zero-3-style sharding across GPUs to reduce overhead. During each step, each GPU updates only its own shard.}
    \label{fig:ema}
\end{wrapfigure}
EMA offers more stable and consistent weight averaging during training and is stored in FP32 to preserve numerical precision. This approach ensures that weight fluctuations are smoothed over time, which improves convergence behavior and promotes better generalization. Because our model is extremely large, storing an extra FP32 copy on each GPU would strain computational resources and potentially limit overall batch size. As shown in Figure~\ref{fig:ema}, the training model (DiT) operates under ZeRO-2, while the EMA model is sharded across GPUs using ZeRO-3. By leveraging ZeRO-3 for the EMA, each GPU holds only a fraction of the full FP32 parameters, enabling efficient memory utilization. For instance, a 20B model sharded across $N$ GPUs requires each GPU to hold only $\frac{20 \times 4}{N}$ GiB, which minimizes redundant storage. We update the EMA every step, which reduces its computation to $\frac{1}{N}$ and ensures that computation cost remains low as the number of GPUs increases. This scheme also supports running the training model under ZeRO-3, further decreasing memory overhead and allowing larger effective batch sizes.

\subsection{Training Data}

We use almost identical data in two stages. This includes open‐source high‐quality data, self‐generated data, and filtered open‐source data. Data types include:
\begin{enumerate}
  \item \textbf{Image Perception:} canny, mlsd, hed, depth, sketch, segmentation (mask), detection (bounding box). Most data originates from (1) Graph200k~\cite{li2025visualcloze}, (2) COCO2017~\cite{lin2014microsoft}. Although perception maps usually reach $1024\times1024$ resolution, over 90\% of reference images are limited to lower resolutions (e.g., $512\times512$). Since perception maps and reference images differ substantially, the mask weighting strategy is unnecessary. Image perception data amounts to approximately 1.4M.
  
  \item \textbf{Image Manipulation:} common edit types such as add, remove, replace, etc. Main sources are ImgEdit~\cite{imgedit} and SEED-X~\cite{SEED-X}. ImgEdit provides over 1M editing samples. We use the subset with high score, totaling 724k higher-quality samples. In SEED-X, we select part 3 with resolutions of at least $1024\times1024$. We also collect style transfers from Graph200k given reference style images, as well as virtual try-on and product extraction data. Since most open-source data lack editing masks, we generate edit masks following Section~\ref{sec:weight}. Image manipulation data amounts to approximately 1M.
  
  \item \textbf{Text-to-Image Generation:} sources include BLIP3-o~\cite{BLIP3-o} and internal images from the Open-Sora Plan~\cite{lin2024open}. Open-Sora Plan images receive dense captions from Qwen2-VL-72B~\cite{Qwen2VL}, all with resolutions of at least $1024\times1024$ and aesthetic score at least 6.0. Text-to-image generation data amounts to approximately 300k.
\end{enumerate}

\begin{figure*}[h]
    \centering
    \includegraphics[width=1\linewidth]{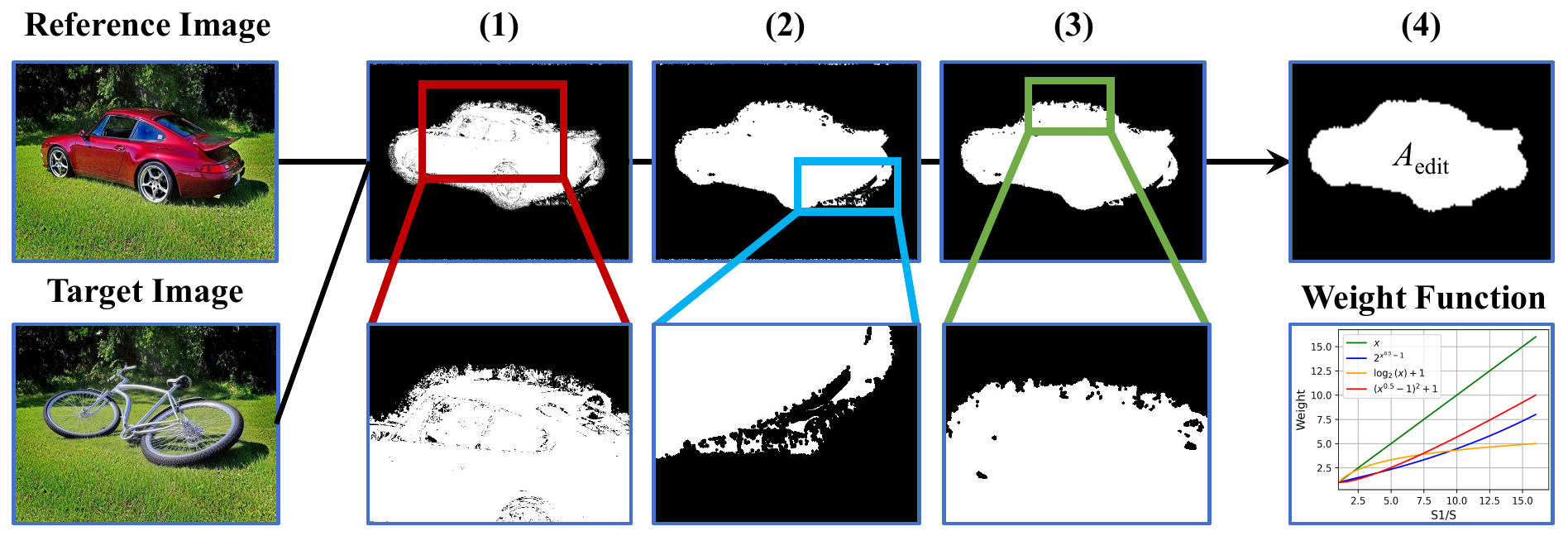}
    \caption{\textbf{Pipeline for mask generation.} Given a reference image and a target image, the mask is obtained through (1) pixel-wise differencing, (2) dilation, (3) connected component filtering, and (4) max-pooling downsampling. The bottom right shows four different weighting functions. We highlight the limitations of steps (1), (2), and (3), which are addressed in the next stage.
    }
    \label{fig:weight}
\end{figure*}

\subsection{Adaptive Editing Region Weighting Strategy}
\label{sec:weight}
In image editing tasks, the edited region typically occupies only a small portion of the image. If uniform loss weighting is applied across the whole image, the loss signal from the relatively small edited area may be overwhelmed by that from the much larger unedited region. This imbalance can cause the model to underfit the edited content, failing to capture fine-grained or user-intended changes.

A straightforward approach is to use the mask region (i.e., the edited area) as a weight. However, not all edited data comes with masks. Therefore, we obtain masks through the following four steps:

(1) \textbf{Pixel-wise Differencing:} Compute the pixel-wise difference between the reference and target images. We set a tolerance threshold to determine whether a pixel region is edited or unedited. However, as shown in Figure~\ref{fig:weight} (1), this produces many noisy differences.

(2) \textbf{Dilation:} Expand each edited pixel using a dilation factor to reduce noise, though many isolated pixels may still remain.

(3) \textbf{Connected Component Filtering:} Remove small connected components to eliminate spurious edits, but this does not address bubbles within larger edited regions.

(4) \textbf{Max-pooling Downsampling:} Apply max-pooling to remove internal noise within connected regions.
Finally, we obtain the edited area size, denoted as $A_{\text{edit}}$.
 
To prevent small edited regions from being overwhelmed by the background during training, we design a weighting strategy that assigns higher loss weights to edited pixels. The weight is a function of the relative area ratio between the full image and the edited region: $x = \frac{A_{\text{total}}}{A_{\text{edit}}}$, where the total image area is denoted as $A_{\text{total}}$. We require the weighting function $w(x)$ to satisfy $w(1) = 1$, so that when the entire image is edited (i.e., text-to-image or style transfer, $A_{\text{edit}} = A_{\text{total}}$), the loss reverts to uniform weighting. We design and compare four candidate functions:

\[
\begin{tabular}{@{} l l @{}}
(1)\quad Linear:           & $w(x) = x$ \\[0.5em]
(2)\quad Exponential Root:  & $w(x) = 2^{\sqrt{x} - 1}$ \\[0.5em]
(3)\quad Logarithmic:       & $w(x) = \log_2(x) + 1$ \\[0.5em]
(4)\quad Quadratic Root:    & $w(x) = (\sqrt{x} - 1)^2 + 1$
\end{tabular}
\]

All four functions ensure $w(1) = 1$, and increase as the edit area shrinks ($x \to \infty$). Among them, the logarithmic function (3) grows moderately, avoids instability for extremely small regions, and maintains a good balance between sensitivity and robustness. Therefore, we adopt the logarithmic weighting function $w(x) = \log_2(x) + 1$ in our final implementation.

\section{Experiment}\label{sec: experiment}

\subsection{Main Results}
 Table \ref{tab:main_res} presents our main comparative results. We compare the performance of our UniWorld-V1 model against other advanced models across three core benchmarks: image understanding, image generation, and image editing. The experimental results comprehensively demonstrate UniWorld-V1's exceptional performance across these three major categories, proving its powerful unified capabilities. We achieved state-of-the-art or near state-of-the-art performance in every sub-category. 
 
Specific comparisons for text-to-image generation will be provided in Section \ref{subsec: t2i}, image editing results in Section \ref{subsec: Manipulation}, visual understanding comparisons in Section \ref{subsec: Understanding}, and image perception capabilities will be showcased through sampling examples in Section \ref{subsec: Perception}.

\begin{table*}[t]
\scriptsize
  \setlength\tabcolsep{0.75mm}
  \caption{\textbf{Comparison between different models on Understanding \& Generation \& Editing benchmarks.} $\dag$ refer to the methods using LLM rewriter. ``×'' indicates the model is incapable of performing the task.}
  \label{tab:main_res}
  \centering
  \begin{tabular}{l|cccc|cc|ccccccc}
    \toprule
     \multirow{2}{*}{\textbf{Model}} & \multicolumn{4}{c|}{\textbf{Understanding}} & \multicolumn{2}{c|}{\textbf{Image Generation}} & \multicolumn{7}{c}{\textbf{Image Editing}} \\
      & MMB$^V$ & MMB$^I$ & MMMU & MM-Vet  & GenEval & WISE  & Overall &  Add &  Adjust &  Extract  &  Replace & Remove  & Hybird \\
    \midrule
    \multicolumn{14}{c}{\textbf{\textit{Image Understanding}}} \\
    % \midrule
    LLaVA-1.5~\cite{LLaVA-1.5} & × & 36.4  & 67.8  &  36.3  &  × & ×  & × & ×  &  × &  ×&  × &  ×  &  × \\
    LLaVA-NeXT~\cite{LLaVA-NeXT}&  × & 79.3  & 51.1  &  57.4  &  × & ×  & × & ×  &  × &  ×&  × &  × &  ×  \\
    \midrule
    \multicolumn{13}{c}{\textbf{\textit{Image \& Video Understanding}}} \\
    % \midrule
    Video-LLaVA~\cite{Video-LLaVA} & \underline{1.05} &  60.9 &  32.8 & 32.0    &  × & ×  & × & ×  &  × &  × &  × &  × &  × \\
    LLaVA-OV~\cite{LLaVA-OV} & 0.94 &  80.8 & 48.8  & 57.5   &  × & ×  & × & ×  &  × &  ×&  × &  ×   &  ×\\
    \midrule
    \multicolumn{14}{c}{\textbf{\textit{Text-to-Image Generation}}} \\
    % \midrule
    SDXL~\cite{SDXL}&  × &  × &  × &  ×  & 0.55  &  0.55  & × &  × &  × &  × &  × &  ×  &  ×\\
    FLUX.1 Dev~\cite{FLUX}&  × &  × &  × &  ×  & 0.66  & 0.50   & × &  × &  × &  ×&  × &  ×  &  × \\
    \midrule
    \multicolumn{14}{c}{\textbf{\textit{Image Editing}}} \\
    % \midrule
    % MagicBrush~\cite{MagicBrush} & × &  × &  × &  ×  & ×  & ×   & 1.85 &  2.72 & 1.47  &  1.31 & 1.89  & 1.57 & 1.80 \\
    % Instruct-P2P~\cite{Instruct-P2P} & × &  × &  ×  &  × & ×  & ×  & 1.89 & 2.29 &  1.79 &  1.33 & 1.93  & 1.49  & 1.48 \\
    % AnyEdit~\cite{AnyEdit} & ×  &  × &  × &  × & ×  & ×  & 2.63 &  3.12 &  2.66 &  1.82&  2.71 &   2.34  & 2.07  \\
    % Step1-Edit~\cite{Step1-Edit}&  × &  × &  × &  × & ×  & ×  & \underline{3.17} &  \textbf{3.90} & 3.13 &  \underline{1.87} &  \underline{3.45} &  \underline{2.61} &2.52 \\
MagicBrush~\cite{MagicBrush} & × &  × &  × &  ×  & ×  & ×   & 1.83 & 2.84 & 1.58 & 1.51 & 1.97 & 1.58 & 1.62 \\
Instruct-P2P~\cite{Instruct-P2P} & × &  × &  ×  &  × & ×  & ×  & 1.88 & 2.45 & 1.83 & 1.44 & 2.01 & 1.50 & 1.20 \\
AnyEdit~\cite{AnyEdit} & ×  &  × &  × &  × & ×  & ×  & 2.45 & 3.18 & 2.95 & 1.88 & 2.47 & 2.23 & 1.56 \\
UltraEdit~\cite{UltraEdit}&  × &  × &  × &  × & ×  & ×  & 2.70 & 3.44 & 2.81 & \underline{2.13} & 2.96 & 1.45 & 1.91 \\
Step1X-Edit~\cite{Step1-Edit}&  × &  × &  × &  × & ×  & ×  & 3.06 & \textbf{3.88} & 3.14 & 1.76 & \underline{3.40} & 2.41 & \underline{2.64} \\
    
    \midrule
    \multicolumn{14}{c}{\textbf{\textit{Unified Understanding \& Generation}}} \\
    % \midrule
    Show-o~\cite{Show-o} & × & -  & 27.4  & -   &  0.68 & 0.35  &  × &  × &  × &  × &  × &  ×   &  × \\
    Janus~\cite{Janus} & × &  69.4 &  30.5 &  34.3  & 0.61  & 0.18 &  × & × & ×  & × &  × &  ×   &  ×  \\
    Janus-Pro~\cite{Janus-Pro} & × &  75.5 &  36.3 & 39.8  & 0.80  & 0.35    & ×  &  × & ×  &  × &  ×&  ×   & ×   \\
    Emu3~\cite{Emu3} & - & 58.5 &31.6  &  37.2 & 0.66$^\dag$  & 0.39   & -  & - & - &   - &  - &  -  &  - \\

    MetaQuery-XL~\cite{MetaQuery} & - & 83.5  & \textbf{58.6}  &  66.6  &  0.80$^\dag$ & \textbf{0.55}   & -  & - &  - &  - &  - &  -   &  - \\
BAGEL~\cite{BAGEL} & - & \textbf{85.0} & \underline{55.3} &  \textbf{67.2}  &  \textbf{0.88$^\dag$} &  \underline{0.52}  & \underline{3.20} & 3.56 & \underline{3.31} & 1.70 & 3.30 & \underline{2.62} & 2.38 \\
    \rowcolor{myblue}
\textbf{UniWorld-V1} & \textbf{1.79} & \underline{83.5} & \textbf{58.6} & \underline{67.1}  & \underline{0.84}$^\dag$ & \textbf{0.55}  & \textbf{3.26} & \underline{3.82} & \textbf{3.64} & \textbf{2.27} & \textbf{3.47} & \textbf{3.24} & \textbf{2.96} \\

    \bottomrule
  \end{tabular}
    \vspace{-12pt}

\end{table*}

\subsection{Text-to-Image Generation}
\label{subsec: t2i}

This section presents the performance of our UniWorld-V1 in text-to-image generation, primarily focusing on two aspects: the models' fundamental object-focused text-to-image generation abilities, as evaluated by the GenEval~\cite{ghosh2023geneval}, and their world knowledge reasoning capabilities for image generation, as assessed by the WISE~\cite{niu2025wise}.

\noindent{\textbf{Evaluation results on GenEval.}} Table \ref{tab:geneval} showcases the models' evaluation results on GenEval. Our UniWorld-V1 achieves a strong performance with an overall score of 0.79. Furthermore, we observe that many models, such as MetaQuery, BLIP3-o, and BAGEL, utilize LLM rewriters for prompt rewriting to facilitate generation. For a fair comparison, we used the rewrite prompt used by blip3o for additional testing. UniWorld-V1 ultimately achieves an impressive score of 0.84. This result is remarkably close to the top-performing BAGEL's 0.88, while UniWorld-V1 only utilizes 2.7M training data compared to BAGEL's 2665M data. This stark difference in data efficiency powerfully highlights the superiority of our architecture.

\noindent{\textbf{Evaluation results on WISE.}} 
As presented in Table \ref{tab:wisescore}, our proposed UniWorld-V1 exhibits exceptionally strong performance as a unified model, achieving an overall score of 0.55. This makes UniWorld-V1 highly competitive among unified models in leveraging and integrating world knowledge for generating semantically rich and accurate images. Notably, UniWorld-V1 achieves an impressive 0.73 in the Space category. This score is particularly significant as it is the closest to GPT-4o-Image's 0.89 in this category, positioning UniWorld-V1 as the top performer among all other evaluated models (excluding GPT-4o-Image) in capturing and utilizing spatial world knowledge.

\begin{table*}[t]
    \small
    \centering
    \setlength\tabcolsep{0.6mm}
    \caption{\textbf{Comparison results on GenEval.} “Gen. Only” refers to pure generation models, while “Unified” indicates models capable of both understanding and generation. $\dagger$ refers to the methods using LLM rewriter. $\ddag$: Results of GPT-4o-Image are tested by~\cite{gpt_imgeval}.
    }
    \begin{tabular}{l|cccccc|c}
        \toprule
        \textbf{Model}  & \textbf{Single Obj.$\uparrow$} & \textbf{Two Obj.$\uparrow$} & \textbf{Counting$\uparrow$} & \textbf{Colors$\uparrow$} & \textbf{Position$\uparrow$} & \textbf{Color Attribute$\uparrow$} & \textbf{Overall$\uparrow$} \\
        \midrule
        \multicolumn{8}{c}{\textbf{\textit{Gen. Only}}} \\
        PixArt-$\alpha$~\cite{PixArt} &  0.98 & 0.50 & 0.44 & 0.80 & 0.08 & 0.07 & 0.48 \\
        Emu$3$-Gen ~\cite{Emu3}  & 0.98 & 0.71 & 0.34 & 0.81 & 0.17 & 0.21 & 0.54 \\
        SDXL~\cite{SDXL} &  0.98 & 0.74 & 0.39 & 0.85 & 0.15 & 0.23 & 0.55 \\
        DALL-E $3$~\cite{DALLE3} & 0.96 & 0.87 & 0.47 & 0.83 & 0.43 & 0.45 & 0.67 \\
        SD3-Medium~\cite{SD3} & 0.99 & 0.94 & 0.72 & 0.89 & 0.33 & 0.60 & 0.74 \\
        FLUX.1-dev$^{\dagger}$~\cite{FLUX} & 0.98 & 0.93 & 0.75 & 0.93 & 0.68 & 0.65 & \emph{0.82} \\
        \midrule
        \multicolumn{8}{c}{\textbf{\textit{Unified}}} \\
        Janus~\cite{Janus} & 0.97 & 0.68 & 0.30 & 0.84 & 0.46 & 0.42 & 0.61 \\
        Emu$3$-Gen$^{\dagger}$\cite{Emu3} & 0.99 & 0.81 & 0.42 & 0.80 & 0.49 & 0.45 & 0.66 \\
        Show-o~\cite{Show-o} &  0.98 & 0.80 & 0.66 & 0.84 & 0.31 & 0.50 & 0.68 \\
        Janus-Pro-7B~\cite{Janus-Pro} &  0.99 & 0.89 & 0.59 & 0.90 & 0.79 & 0.66 & 0.80 \\
        MetaQuery-XL$^{\dagger}$~\cite{MetaQuery} &  -& - & - & -& -& -& 0.80 \\
        BLIP3-o~\cite{BLIP3-o} &  -& - & - & -& -& -& 0.84 \\
        BAGEL~\cite{BAGEL} & 0.99 & 0.94  & 0.81 & 0.88 & 0.64 &0.63 & \emph{0.82}\\
        BAGEL$^{\dagger}$~\cite{BAGEL} & 0.98 & 0.95  & 0.84 & 0.95 & 0.78 &0.77 & \textbf{0.88} \\
        GPT-4o-Image$^\ddag$ & 0.99 & 0.92 & 0.85 & 0.92 & 0.75 & 0.61 & \underline{0.84} \\
        \rowcolor{myblue}
        \textbf{UniWorld-V1} & 0.99  & 0.93 & 0.79 & 0.89 & 0.49 & 0.70 & 0.80 \\
        \rowcolor{myblue}
        \textbf{UniWorld-V1}$^{\dagger}$ & 0.98  & 0.93 & 0.81 & 0.89 & 0.74 & 0.71 & \underline{0.84} \\
    \bottomrule
    \end{tabular}
    \label{tab:geneval}
\end{table*}

\begin{table*}[!t]
    \small
    \centering
    \setlength\tabcolsep{1.9mm}
    \caption{\textbf{Comparison results on WISE.} WISE evaluates the model's capacity for complex semantic understanding and world knowledge in text-to-image generation. “Gen. Only” refers to pure generation models, while “Unified” indicates models capable of both understanding and generation. $^\dag$: Results of GPT-4o-Image are tested by~\cite{gpt_imgeval}.
    }
    \label{tab:wisescore}
    \begin{tabular}{l|cccccc|c}
    \toprule
    \textbf{Model} & \textbf{Cultural$\uparrow$}  & \textbf{Time$\uparrow$}     & \textbf{Space$\uparrow$}    & \textbf{Biology$\uparrow$}    & \textbf{Physics$\uparrow$} & \textbf{Chemistry$\uparrow$} & \textbf{Overall$\uparrow$} \\
    \midrule
    
    \multicolumn{8}{c}{\textbf{\textit{Gen. Only}}} \\
    SDXL~\cite{SDXL} &0.43  & 0.48 &0.47  &0.44  &0.45 &0.27 & 0.43 \\
    SD3.5-large~\cite{SD3} & 0.44 &0.50 &0.58  & 0.44&0.52 &0.31 & 0.46 \\
    PixArt-Alpha~\cite{PixArt} & 0.45  & 0.50& 0.48 & 0.49& 0.56 &0.34 & 0.47\\
    playground-v2.5~\cite{playground} & 0.49  &0.58  & 0.55&0.43  & 0.48&0.33 & 0.49 \\
    FLUX.1-dev~\cite{FLUX} & 0.48  &0.58 &0.62  &0.42  &0.51 & 0.35 & 0.50 \\
    \midrule
    \multicolumn{8}{c}{\textbf{\textit{Unified}}} \\
    Janus~\cite{Janus} &0.16 &0.26 &0.35 & 0.28 &0.30 & 0.14& 0.23\\
    Show-o~\cite{Show-o} & 0.28 &0.40  &0.48 & 0.30& 0.46 & 0.30 & 0.35\\
    Janus-Pro-7B~\cite{Janus-Pro} & 0.30& 0.37& 0.49 & 0.36&0.42 &0.26 & 0.35 \\
    Emu3~\cite{Emu3} & 0.34 & 0.45 & 0.48 & 0.41  & 0.45 & 0.27 & 0.39 \\
    MetaQuery-XL~\cite{MetaQuery} & 0.56& 0.55 &0.62 &  0.49 &  0.63 & 0.41 & \underline{0.55} \\
    BAGEL~\cite{BAGEL} & 0.44 & 0.55 & 0.68 & 0.44 & 0.60 & 0.39 & 0.52 \\
    GPT-4o-Image$^\dag$ &0.81 &0.71 &0.89 &0.83 &0.79 &0.74 & \textbf{0.80} \\
    \rowcolor{myblue}
    \textbf{UniWorld-V1} & 0.53  & 0.55   & 0.73 &0.45  &0.59  & 0.41  & \underline{0.55} \\
    \bottomrule
    \end{tabular}
\end{table*}

\subsection{Image Manipulation}
\label{subsec: Manipulation}
As illustrated in Table~\ref{tab:imgeditscore}, we compare UniWorld-V1 with other open-source models and GPT-4o-Image on editing capabilities using the \textit{ImgEdit-Bench}~\cite{imgedit} benchmark. Our UniWorld-V1 model achieves the best overall performance among all open-source models, with a total score of 3.26, outperforming other strong open-source competitors such as \textit{BAGEL} (3.20) and \textit{Step1X-Edit} (3.06). This highlights UniWorld-V1’s robust and consistent image editing ability across a wide spectrum of tasks. Notably, UniWorld-V1 ranks first among open-source models in several key categories, including \textit{Adjust} (3.64), \textit{Extract} (2.27), \textit{Replace} (3.47), \textit{Remove} (3.24), \textit{Background} (2.99), and \textit{Hybrid} (2.96). These results reflect UniWorld-V1’s strong capability in tasks that require nuanced attribute adjustment, structural modification, and complex multi-operation editing. While \textit{GPT-4o-Image} continues to lead with a remarkable total score of 4.20, UniWorld-V1 emerges as the \textit{closest open-source contender}, significantly narrowing the performance gap. This demonstrates UniWorld-V1’s notable progress towards achieving high-fidelity and generalizable image editing quality, comparable to state-of-the-art proprietary models.

\begin{table*}[t]
    \small
    \centering
    \setlength\tabcolsep{0.8mm}
    \caption{\textbf{Comparison results on ImgEdit-Bench.} ``Overall'' is calculated by averaging all scores across tasks. We use GPT-4.1 for evaluation.}
    \label{tab:imgeditscore}
    \begin{tabular}{l|ccccccccc|c}
    \toprule
    \textbf{Model} & \textbf{Add} & \textbf{Adjust} & \textbf{Extract} & \textbf{Replace} & \textbf{Remove} & \textbf{Background} & \textbf{Style} & \textbf{Hybrid} & \textbf{Action} & \textbf{Overall$\uparrow$} \\
    \midrule

    MagicBrush~\cite{MagicBrush} & 2.84 & 1.58 & 1.51 & 1.97 & 1.58 & 1.75 & 2.38 & 1.62 & 1.22 & 1.83 \\

    Instruct-P2P~\cite{Instruct-P2P} & 2.45 & 1.83 & 1.44 & 2.01 & 1.50 & 1.44 & 3.55 & 1.20 & 1.46 & 1.88 \\

    AnyEdit~\cite{AnyEdit} & 3.18 & 2.95 & 1.88 & 2.47 & 2.23 & 2.24 & 2.85 & 1.56 & 2.65 & 2.45 \\

    UltraEdit~\cite{UltraEdit} & 3.44 & 2.81 & 2.13 & 2.96 & 1.45 & 2.83 & 3.76 & 1.91 & 2.98 & 2.70 \\

    Step1X-Edit~\cite{Step1-Edit} & \underline{3.88} & 3.14 & 1.76 & 3.40 & 2.41 & 3.16 & \underline{4.63} & 2.64 & 2.52 & 3.06 \\

    BAGEL~\cite{BAGEL} & 3.56 & 3.31 & 1.70 & 3.30 & 2.62 & 3.24 & 4.49 & 2.38 & \underline{4.17} & 3.20 \\

    GPT-4o-Image & \textbf{4.61} & \textbf{4.33} & \textbf{2.90} & \textbf{4.35} & \textbf{3.66} & \textbf{4.57} & \textbf{4.93} & \textbf{3.96} & \textbf{4.89} & \textbf{4.20} \\

    \rowcolor{myblue}
    \textbf{UniWorld-V1} & 3.82 & \underline{3.64} & \underline{2.27} & \underline{3.47} & \underline{3.24} & \underline{2.99} & 4.21 & \underline{2.96} & 2.74 & \underline{3.26} \\

    \bottomrule
    \end{tabular}
\end{table*}

\begin{wraptable}{r}{0.5\textwidth} 
  \vspace{-12pt}
    \centering
    % \small
    \setlength\tabcolsep{2.0mm}
    \caption{\textbf{Comparison results on GEdit-Bench.} G\_SC, G\_PQ, and G\_O refer to the metrics evaluated by GPT-4.1.}
    \begin{tabular}{l|ccc}
    \toprule
    \textbf{Model} & \textbf{G\_SC}$\uparrow$ & \textbf{G\_PQ}$\uparrow$ & \textbf{G\_O}$\uparrow$  \\
    \midrule
    Instruct-P2P~\cite{Instruct-P2P} & 3.58 & 5.49 & 3.68 \\
    MagicBrush~\cite{MagicBrush}  & 4.68 & 5.66 & 4.52 \\
    AnyEdit~\cite{AnyEdit}  & 3.18 & 5.82 & 3.21 \\
    OmniGen~\cite{xiao2024omnigen} & 5.96 & 5.89 & 5.06 \\
    Step1X-Edit~\cite{step1x_edit}& 7.09 & 6.76 & \underline{6.70}  \\
    BAGEL~\cite{BAGEL}  & \underline{7.36} & {6.83}& 6.52 \\
    Gemini 2.0~\cite{gemini220250312}  & 6.73 & 6.61 & 6.32 \\
    GPT-4o~\cite{openai2025chatgpt4o} & \textbf{7.85} & \textbf{7.62} & \textbf{7.53}  \\
    \rowcolor{myblue}
    \textbf{UniWorld-V1}  & 4.93 & \underline{7.43}  &  4.85 \\
    \bottomrule
    \end{tabular}
\label{tab:gedit}
\end{wraptable}

To validate out-of-domain generalization, we use GEdit-Bench~\cite{Step1-Edit} and report scores on the English dataset as shown in Table~\ref{tab:gedit}. UniWorld-V1’s perceptual score on GEdit-Bench (G\_PQ) is relatively high (7.43), whereas its instruction-following score (G\_SC) is relatively low (4.93). This is because most of our editing data originates from ImgEdit~\cite{imgedit} and is limited in quantity, resulting in lower instruction diversity than BAGEL or Step1X-Edit. Additionally, text editing is a key evaluation metric in GEdit-Bench. Our dataset contains no text-editing samples, so it performs poorly in this aspect. Moreover, text editing is challenging for a 512-resolution SigLIP. We believe instruction-following capability can be improved by incorporating more data and fine-tuning the VLM. Regarding text-editing capability, we continue collecting relevant data and raising SigLIP’s resolution. Finally, we reiterate that our core contribution is finding that high-resolution SigLIP enables control over reference-image consistency.

\begin{wraptable}{r}{0.5\textwidth} 
  \vspace{-12pt}
  \centering
  \scriptsize
  \setlength\tabcolsep{0.12mm}
  \caption{\textbf{Comparison between different models on Visual Understanding benchmarks.}  × indicates the model is incapable of performing the task.}
  \label{tab:image_res_wrapped}
  \begin{tabular}{l|cccc}
    \toprule
     \textbf{Model} & MMB$^V$~\cite{fang2024mmbench} & MMB$^I$~\cite{liu2024mmbench} & MMMU~\cite{yue2024mmmu} & MM-Vet~\cite{yu2023mm}  \\
    \midrule
    % \multicolumn{5}{c}{\textbf{\textit{Image Understanding}}} \\
    \multicolumn{5}{c}{\textbf{\textit{Image \& Video Understanding}}} \\
    LLaVA-1.5~\cite{LLaVA-1.5} & × & 36.4  & 67.8  &  36.3  \\
    % LLaVA-NeXT~\cite{LLaVA-NeXT} &  × & 79.3  & 51.1  &  57.4  \\
    % \midrule
    % \multicolumn{5}{c}{\textbf{\textit{Image \& Video Understanding}}} \\
    Video-LLaVA~\cite{Video-LLaVA} & 1.05 &  60.9 &  32.8 & 32.0  \\
    % LLaVA-OV~\cite{LLaVA-OV} & 0.94 &  80.8 & 48.8  & 57.5    \\
    \midrule
    \multicolumn{5}{c}{\textbf{\textit{Unified Understanding \& Generation}}} \\
    Show-o~\cite{Show-o} & × & -  & 27.4  & - \\
    Janus~\cite{Janus} & × &  69.4 &  30.5 &  34.3  \\
    Janus-Pro~\cite{Janus-Pro} & × &  75.5 &  36.3 & 39.8  \\
    Emu3~\cite{Emu3} & - & 58.5 &31.6  &  37.2 \\
    BLIP3-o~\cite{BLIP3-o} & - & 83.5  & 50.6  &  66.6   \\
    MetaQuery~\cite{MetaQuery} & - & 83.5  & \textbf{58.6}  &  66.6  \\
    BAGEL~\cite{BAGEL} & - & \textbf{85.0} & 55.3 &  \textbf{67.2}    \\
    GPT-4o & 2.15  &   - & 72.9  & 76.9     \\
    \rowcolor{myblue}
    \textbf{UniWorld-V1} & \textbf{1.79} & 83.5 & \textbf{58.6} & 67.1  \\
    \bottomrule
  \end{tabular}
\end{wraptable}

\vspace{-12pt}
\subsection{Visual Understanding}
\label{subsec: Understanding}

By benefiting from our strategy of freezing the Multimodal Large Language Model component, we successfully inherited the robust multimodal understanding capabilities of Qwen2.5-VL-7B without the need for retraining. This significantly reduces our resource consumption, specifically in terms of data and computational power, and also prevents potential degradation of understanding performance that could arise from training on generative tasks. As presented in Table \ref{tab:image_res_wrapped}, this architectural choice enables UniWorld-V1 to achieve remarkable results, significantly surpassing models like Janus, Show-o, and Emu3  across multiple metrics, and achieves highly competitive performance against the latest advanced models such as BAGEL.

\subsection{Image Perception}
\label{subsec: Perception}
\begin{figure*}[t]
    \centering
    \includegraphics[width=1\linewidth]{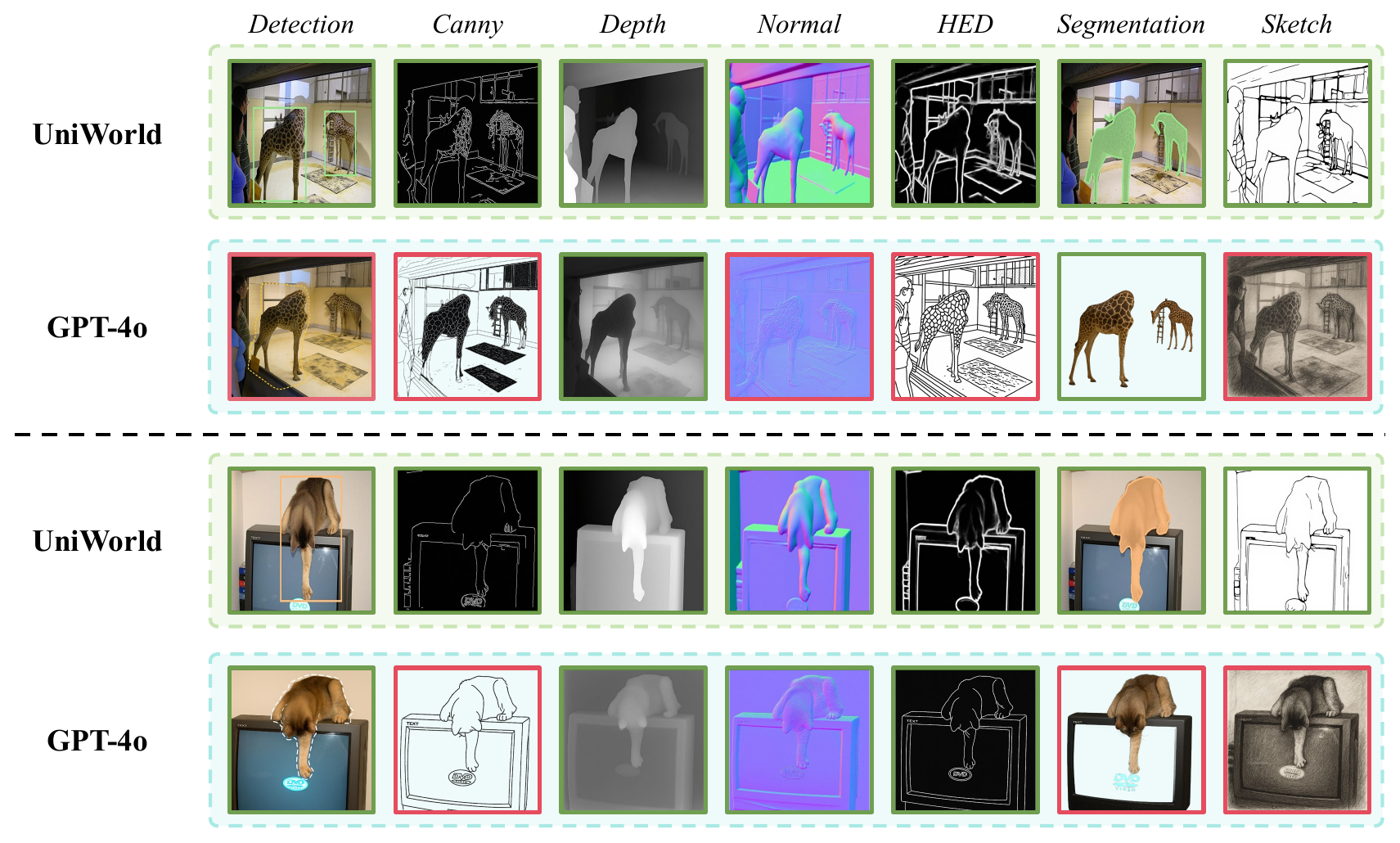}
\caption{\textbf{Showcase of UniWorld-V1’s perception capabilities.} This figure presents a qualitative comparison of UniWorld-V1's perception tasks against GPT-4o, using randomly selected examples. \textcolor{green}{Green boxes} indicate correct responses, while \textcolor{red}{Red boxes} highlight instances where the model's output deviates from the expected result.
    }
    \label{fig:cv}
\end{figure*}

As the first to integrate visual understanding, image perception and manipulation into a unified model, there is currently no suitable benchmark to comprehensively evaluate our UniWorld-V1 model's full image perception capabilities. Therefore, we conducted a qualitative comparison using sampled examples against GPT-4o-Image, as detailed in Figure \ref{fig:cv}. It can be observed that UniWorld-V1 demonstrates highly competitive, and in many scopes superior, performance across various perception tasks. Specifically, UniWorld-V1 distinctly showcases stronger instruction understanding and task execution capabilities than GPT-4o-Image in canny edge detection, normal map generation, HED, segmentation, and sketch generation. This highlights that UniWorld-V1's integrated architecture effectively enables a broad and accurate range of image perception functionalities, positioning it as the first open-source unified model capable of such diverse and high-fidelity visual analyses.

\section{Conclusion}

In summary, UniWorld-V1 demonstrates that a unified architecture anchored by a high-resolution semantic encoder, which address both image perception and manipulation tasks with state-of-the-art efficiency. By leveraging only 2.7M training samples, UniWorld-V1 achieves superior performance against much larger cost models on diverse benchmarks, confirming that semantic encoders provide richer and more versatile visual representations than traditional VAE-based approaches. This work establishes a foundation for future research in unified visual generation. We release all code, model weights, and datasets to foster continued innovation and collaboration within the community.

\textbf{Limitation}
Despite UniWorld-V1’s remarkable training efficiency, the following shortcomings remain:

\begin{itemize}
\item \textbf{Insufficient instruction generalization.} Limited training data and lack of VLM fine-tuning require specific instruction templates to outperform BAGEL.

\item \textbf{Inadequate reference-image consistency.} Reference images are processed at $512\times512$ resolution, which is insufficient to generate all details at $1024\times1024$ scale.

\item \textbf{Incomplete benchmarks.} DPG-Bench and GenAI-Bench with scores above a certain threshold often fail to reflect human preference, as verified through manual inspection. Some samples in GenEval forcibly bind two objects that rarely co-occur in the real world, ignoring the natural distribution of images. ImgEdit-Bench and GEdit-Bench lack sufficient sensitivity to the reference regions.

\end{itemize}

\textbf{Future Work}

\begin{itemize}
\item Continue collecting data and perform joint training with a VLM.

\item Integrate higher-resolution semantic encoders or adopt VLM techniques to increase input-image resolution, such as multi-scale image gridding.

\end{itemize}

\textbf{Failed Attempts}

\begin{itemize}

\item We empirically attempt to replace SigLIP with other encoders such as DINO V2~\cite{oquab2023dinov2} and RADIO V2.5~\cite{RADIOv2p5}, but the attempts are unsuccessful.

\item We attempt to use Qwen2.5VL’s visual output (picking the image feature of outputs while abandoning text feature) directly as the reference-image control signal. However, consistency between generated and reference images remains poor. This issue arises from the intrinsic gap between VLM training objectives and contrastive training. Contrastive learning focuses on global semantic features that saturate as resolution increases, whereas VLM training demands both global and local semantic information. As a result, the model capacity does not preserve sufficient low-level control signals, which likely causes the failure.

\end{itemize}

% \newpage

\bibliographystyle{plain}
\bibliography{ref}

\begin{thebibliography}{10}

\bibitem{gpt4}
Josh Achiam, Steven Adler, Sandhini Agarwal, Lama Ahmad, Ilge Akkaya, Florencia~Leoni Aleman, Diogo Almeida, Janko Altenschmidt, Sam Altman, Shyamal Anadkat, et~al.
\newblock Gpt-4 technical report.
\newblock {\em arXiv preprint arXiv:2303.08774}, 2023.

\bibitem{Qwen25VL}
Shuai Bai, Keqin Chen, Xuejing Liu, Jialin Wang, Wenbin Ge, Sibo Song, Kai Dang, Peng Wang, Shijie Wang, Jun Tang, Humen Zhong, Yuanzhi Zhu, Mingkun Yang, Zhaohai Li, Jianqiang Wan, Pengfei Wang, Wei Ding, Zheren Fu, Yiheng Xu, Jiabo Ye, Xi~Zhang, Tianbao Xie, Zesen Cheng, Hang Zhang, Zhibo Yang, Haiyang Xu, and Junyang Lin.
\newblock Qwen2.5-vl technical report.
\newblock {\em arXiv preprint arXiv:2502.13923}, 2025.

\bibitem{Instruct-P2P}
Tim Brooks, Aleksander Holynski, and Alexei~A Efros.
\newblock Instructpix2pix: Learning to follow image editing instructions.
\newblock In {\em Proceedings of the IEEE/CVF conference on computer vision and pattern recognition}, pages 18392--18402, 2023.

\bibitem{BLIP3-o}
Jiuhai Chen, Zhiyang Xu, Xichen Pan, Yushi Hu, Can Qin, Tom Goldstein, Lifu Huang, Tianyi Zhou, Saining Xie, Silvio Savarese, et~al.
\newblock Blip3-o: A family of fully open unified multimodal models-architecture, training and dataset.
\newblock {\em arXiv preprint arXiv:2505.09568}, 2025.

\bibitem{PixArt}
Junsong Chen, Jincheng Yu, Chongjian Ge, Lewei Yao, Enze Xie, Yue Wu, Zhongdao Wang, James Kwok, Ping Luo, Huchuan Lu, et~al.
\newblock Pixart-$\alpha$: Fast training of diffusion transformer for photorealistic text-to-image synthesis.
\newblock {\em arXiv preprint arXiv:2310.00426}, 2023.

\bibitem{chen2025multimodal}
Liang Chen, Shuai Bai, Wenhao Chai, Weichu Xie, Haozhe Zhao, Leon Vinci, Junyang Lin, and Baobao Chang.
\newblock Multimodal representation alignment for image generation: Text-image interleaved control is easier than you think.
\newblock {\em arXiv preprint arXiv:2502.20172}, 2025.

\bibitem{Janus-Pro}
Xiaokang Chen, Zhiyu Wu, Xingchao Liu, Zizheng Pan, Wen Liu, Zhenda Xie, Xingkai Yu, and Chong Ruan.
\newblock Janus-pro: Unified multimodal understanding and generation with data and model scaling.
\newblock {\em arXiv preprint arXiv:2501.17811}, 2025.

\bibitem{yolo}
Tianheng Cheng, Lin Song, Yixiao Ge, Wenyu Liu, Xinggang Wang, and Ying Shan.
\newblock Yolo-world: Real-time open-vocabulary object detection.
\newblock In {\em Proceedings of the IEEE/CVF Conference on Computer Vision and Pattern Recognition}, pages 16901--16911, 2024.

\bibitem{BAGEL}
Chaorui Deng, Deyao Zhu, Kunchang Li, Chenhui Gou, Feng Li, Zeyu Wang, Shu Zhong, Weihao Yu, Xiaonan Nie, Ziang Song, et~al.
\newblock Emerging properties in unified multimodal pretraining.
\newblock {\em arXiv preprint arXiv:2505.14683}, 2025.

\bibitem{SD3}
Patrick Esser, Sumith Kulal, Andreas Blattmann, Rahim Entezari, Jonas M{\"u}ller, Harry Saini, Yam Levi, Dominik Lorenz, Axel Sauer, Frederic Boesel, et~al.
\newblock Scaling rectified flow transformers for high-resolution image synthesis.
\newblock In {\em Forty-first international conference on machine learning}, 2024.

\bibitem{fang2024mmbench}
Xinyu Fang, Kangrui Mao, Haodong Duan, Xiangyu Zhao, Yining Li, Dahua Lin, and Kai Chen.
\newblock Mmbench-video: A long-form multi-shot benchmark for holistic video understanding.
\newblock {\em Advances in Neural Information Processing Systems}, 37:89098--89124, 2024.

\bibitem{SEED-X}
Yuying Ge, Sijie Zhao, Jinguo Zhu, Yixiao Ge, Kun Yi, Lin Song, Chen Li, Xiaohan Ding, and Ying Shan.
\newblock Seed-x: Multimodal models with unified multi-granularity comprehension and generation.
\newblock {\em arXiv preprint arXiv:2404.14396}, 2024.

\bibitem{gemini220250312}
Google Gemini2.
\newblock Experiment with gemini 2.0 flash native image generation, 2025.

\bibitem{ghosh2023geneval}
Dhruba Ghosh, Hannaneh Hajishirzi, and Ludwig Schmidt.
\newblock Geneval: An object-focused framework for evaluating text-to-image alignment.
\newblock {\em Advances in Neural Information Processing Systems}, 36:52132--52152, 2023.

\bibitem{FLUX-Redux}
Black~Forest Labs.
\newblock Flux.
\newblock \url{https://bfl.ai/announcements/24-11-21-tools}, 2024.

\bibitem{FLUX}
Black~Forest Labs.
\newblock Flux.
\newblock \url{https://github.com/black-forest-labs/flux}, 2024.

\bibitem{LLaVA-OV}
Bo~Li, Yuanhan Zhang, Dong Guo, Renrui Zhang, Feng Li, Hao Zhang, Kaichen Zhang, Peiyuan Zhang, Yanwei Li, Ziwei Liu, et~al.
\newblock Llava-onevision: Easy visual task transfer.
\newblock {\em arXiv preprint arXiv:2408.03326}, 2024.

\bibitem{li2025visualcloze}
Zhong-Yu Li, Ruoyi Du, Juncheng Yan, Le~Zhuo, Zhen Li, Peng Gao, Zhanyu Ma, and Ming-Ming Cheng.
\newblock Visualcloze: A universal image generation framework via visual in-context learning.
\newblock {\em arXiv preprint arXiv:2504.07960}, 2025.

\bibitem{liao2025langbridge}
Jiaqi Liao, Yuwei Niu, Fanqing Meng, Hao Li, Changyao Tian, Yinuo Du, Yuwen Xiong, Dianqi Li, Xizhou Zhu, Li~Yuan, et~al.
\newblock Langbridge: Interpreting image as a combination of language embeddings.
\newblock {\em arXiv preprint arXiv:2503.19404}, 2025.

\bibitem{lin2024open}
Bin Lin, Yunyang Ge, Xinhua Cheng, Zongjian Li, Bin Zhu, Shaodong Wang, Xianyi He, Yang Ye, Shenghai Yuan, Liuhan Chen, et~al.
\newblock Open-sora plan: Open-source large video generation model.
\newblock {\em arXiv preprint arXiv:2412.00131}, 2024.

\bibitem{lin2024moe}
Bin Lin, Zhenyu Tang, Yang Ye, Jiaxi Cui, Bin Zhu, Peng Jin, Jinfa Huang, Junwu Zhang, Yatian Pang, Munan Ning, et~al.
\newblock Moe-llava: Mixture of experts for large vision-language models.
\newblock {\em arXiv preprint arXiv:2401.15947}, 2024.

\bibitem{Video-LLaVA}
Bin Lin, Yang Ye, Bin Zhu, Jiaxi Cui, Munan Ning, Peng Jin, and Li~Yuan.
\newblock Video-llava: Learning united visual representation by alignment before projection.
\newblock {\em arXiv preprint arXiv:2311.10122}, 2023.

\bibitem{lin2014microsoft}
Tsung-Yi Lin, Michael Maire, Serge Belongie, James Hays, Pietro Perona, Deva Ramanan, Piotr Doll{\'a}r, and C~Lawrence Zitnick.
\newblock Microsoft coco: Common objects in context.
\newblock In {\em Computer vision--ECCV 2014: 13th European conference, zurich, Switzerland, September 6-12, 2014, proceedings, part v 13}, pages 740--755. Springer, 2014.

\bibitem{playground}
Bingchen Liu, Ehsan Akhgari, Alexander Visheratin, Aleks Kamko, Linmiao Xu, Shivam Shrirao, Chase Lambert, Joao Souza, Suhail Doshi, and Daiqing Li.
\newblock Playground v3: Improving text-to-image alignment with deep-fusion large language models.
\newblock {\em arXiv preprint arXiv:2409.10695}, 2024.

\bibitem{LLaVA-1.5}
Haotian Liu, Chunyuan Li, Yuheng Li, and Yong~Jae Lee.
\newblock Improved baselines with visual instruction tuning.
\newblock In {\em Proceedings of the IEEE/CVF Conference on Computer Vision and Pattern Recognition}, pages 26296--26306, 2024.

\bibitem{step1x_edit}
Shiyu Liu, Yucheng Han, Peng Xing, Fukun Yin, Rui Wang, Wei Cheng, Jiaqi Liao, Yingming Wang, Honghao Fu, Chunrui Han, et~al.
\newblock Step1x-edit: A practical framework for general image editing.
\newblock {\em arXiv preprint arXiv:2504.17761}, 2025.

\bibitem{Step1-Edit}
Shiyu Liu, Yucheng Han, Peng Xing, Fukun Yin, Rui Wang, Wei Cheng, Jiaqi Liao, Yingming Wang, Honghao Fu, Chunrui Han, et~al.
\newblock Step1x-edit: A practical framework for general image editing.
\newblock {\em arXiv preprint arXiv:2504.17761}, 2025.

\bibitem{liu2024mmbench}
Yuan Liu, Haodong Duan, Yuanhan Zhang, Bo~Li, Songyang Zhang, Wangbo Zhao, Yike Yuan, Jiaqi Wang, Conghui He, Ziwei Liu, et~al.
\newblock Mmbench: Is your multi-modal model an all-around player?
\newblock In {\em European conference on computer vision}, pages 216--233. Springer, 2024.

\bibitem{niu2025wise}
Yuwei Niu, Munan Ning, Mengren Zheng, Bin Lin, Peng Jin, Jiaqi Liao, Kunpeng Ning, Bin Zhu, and Li~Yuan.
\newblock Wise: A world knowledge-informed semantic evaluation for text-to-image generation.
\newblock {\em arXiv preprint arXiv:2503.07265}, 2025.

\bibitem{openai2025chatgpt4o}
OpenAI.
\newblock Introducing 4o image generation, 2025.

\bibitem{oquab2023dinov2}
Maxime Oquab, Timoth{\'e}e Darcet, Th{\'e}o Moutakanni, Huy Vo, Marc Szafraniec, Vasil Khalidov, Pierre Fernandez, Daniel Haziza, Francisco Massa, Alaaeldin El-Nouby, et~al.
\newblock Dinov2: Learning robust visual features without supervision.
\newblock {\em arXiv preprint arXiv:2304.07193}, 2023.

\bibitem{MetaQuery}
Xichen Pan, Satya~Narayan Shukla, Aashu Singh, Zhuokai Zhao, Shlok~Kumar Mishra, Jialiang Wang, Zhiyang Xu, Jiuhai Chen, Kunpeng Li, Felix Juefei-Xu, et~al.
\newblock Transfer between modalities with metaqueries.
\newblock {\em arXiv preprint arXiv:2504.06256}, 2025.

\bibitem{peebles2023scalable}
William Peebles and Saining Xie.
\newblock Scalable diffusion models with transformers.
\newblock In {\em Proceedings of the IEEE/CVF international conference on computer vision}, pages 4195--4205, 2023.

\bibitem{SDXL}
Dustin Podell, Zion English, Kyle Lacey, Andreas Blattmann, Tim Dockhorn, Jonas M{\"u}ller, Joe Penna, and Robin Rombach.
\newblock Sdxl: Improving latent diffusion models for high-resolution image synthesis.
\newblock {\em arXiv preprint arXiv:2307.01952}, 2023.

\bibitem{raffel2020exploring}
Colin Raffel, Noam Shazeer, Adam Roberts, Katherine Lee, Sharan Narang, Michael Matena, Yanqi Zhou, Wei Li, and Peter~J Liu.
\newblock Exploring the limits of transfer learning with a unified text-to-text transformer.
\newblock {\em Journal of machine learning research}, 21(140):1--67, 2020.

\bibitem{sam2}
Nikhila Ravi, Valentin Gabeur, Yuan-Ting Hu, Ronghang Hu, Chaitanya Ryali, Tengyu Ma, Haitham Khedr, Roman R{\"a}dle, Chloe Rolland, Laura Gustafson, et~al.
\newblock Sam 2: Segment anything in images and videos.
\newblock {\em arXiv preprint arXiv:2408.00714}, 2024.

\bibitem{DALLE3}
Zhan Shi, Xu~Zhou, Xipeng Qiu, and Xiaodan Zhu.
\newblock Improving image captioning with better use of captions.
\newblock {\em arXiv preprint arXiv:2006.11807}, 2020.

\bibitem{OmniConsistency}
Yiren Song, Cheng Liu, and Mike~Zheng Shou.
\newblock Omniconsistency: Learning style-agnostic consistency from paired stylization data.
\newblock 2025.

\bibitem{RADIOv2p5}
RADIOv2.5 Team.
\newblock Flux.
\newblock \url{https://github.com/NVlabs/RADIO/blob/main/RADIOv2.5_tech_report.md}, 2024.

\bibitem{tong2024metamorph}
Shengbang Tong, David Fan, Jiachen Zhu, Yunyang Xiong, Xinlei Chen, Koustuv Sinha, Michael Rabbat, Yann LeCun, Saining Xie, and Zhuang Liu.
\newblock Metamorph: Multimodal understanding and generation via instruction tuning.
\newblock {\em arXiv preprint arXiv:2412.14164}, 2024.

\bibitem{tschannen2025siglip}
Michael Tschannen, Alexey Gritsenko, Xiao Wang, Muhammad~Ferjad Naeem, Ibrahim Alabdulmohsin, Nikhil Parthasarathy, Talfan Evans, Lucas Beyer, Ye~Xia, Basil Mustafa, et~al.
\newblock Siglip 2: Multilingual vision-language encoders with improved semantic understanding, localization, and dense features.
\newblock {\em arXiv preprint arXiv:2502.14786}, 2025.

\bibitem{Qwen2VL}
Peng Wang, Shuai Bai, Sinan Tan, Shijie Wang, Zhihao Fan, Jinze Bai, Keqin Chen, Xuejing Liu, Jialin Wang, Wenbin Ge, Yang Fan, Kai Dang, Mengfei Du, Xuancheng Ren, Rui Men, Dayiheng Liu, Chang Zhou, Jingren Zhou, and Junyang Lin.
\newblock Qwen2-vl: Enhancing vision-language model's perception of the world at any resolution.
\newblock {\em arXiv preprint arXiv:2409.12191}, 2024.

\bibitem{Emu3}
Xinlong Wang, Xiaosong Zhang, Zhengxiong Luo, Quan Sun, Yufeng Cui, Jinsheng Wang, Fan Zhang, Yueze Wang, Zhen Li, Qiying Yu, et~al.
\newblock Emu3: Next-token prediction is all you need.
\newblock {\em arXiv preprint arXiv:2409.18869}, 2024.

\bibitem{Janus}
Chengyue Wu, Xiaokang Chen, Zhiyu Wu, Yiyang Ma, Xingchao Liu, Zizheng Pan, Wen Liu, Zhenda Xie, Xingkai Yu, Chong Ruan, et~al.
\newblock Janus: Decoupling visual encoding for unified multimodal understanding and generation.
\newblock {\em arXiv preprint arXiv:2410.13848}, 2024.

\bibitem{xiao2024omnigen}
Shitao Xiao, Yueze Wang, Junjie Zhou, Huaying Yuan, Xingrun Xing, Ruiran Yan, Chaofan Li, Shuting Wang, Tiejun Huang, and Zheng Liu.
\newblock Omnigen: Unified image generation.
\newblock {\em arXiv preprint arXiv:2409.11340}, 2024.

\bibitem{Show-o}
Jinheng Xie, Weijia Mao, Zechen Bai, David~Junhao Zhang, Weihao Wang, Kevin~Qinghong Lin, Yuchao Gu, Zhijie Chen, Zhenheng Yang, and Mike~Zheng Shou.
\newblock Show-o: One single transformer to unify multimodal understanding and generation.
\newblock {\em arXiv preprint arXiv:2408.12528}, 2024.

\bibitem{gpt_imgeval}
Zhiyuan Yan, Junyan Ye, Weijia Li, Zilong Huang, Shenghai Yuan, Xiangyang He, Kaiqing Lin, Jun He, Conghui He, and Li~Yuan.
\newblock Gpt-imgeval: A comprehensive benchmark for diagnosing gpt4o in image generation.
\newblock {\em arXiv preprint arXiv:2504.02782}, 2025.

\bibitem{imgedit}
Yang Ye, Xianyi He, Zongjian Li, Bin Lin, Shenghai Yuan, Zhiyuan Yan, Bohan Hou, and Li~Yuan.
\newblock Imgedit: A unified image editing dataset and benchmark.
\newblock {\em arXiv preprint arXiv:2505.20275}, 2025.

\bibitem{AnyEdit}
Qifan Yu, Wei Chow, Zhongqi Yue, Kaihang Pan, Yang Wu, Xiaoyang Wan, Juncheng Li, Siliang Tang, Hanwang Zhang, and Yueting Zhuang.
\newblock Anyedit: Mastering unified high-quality image editing for any idea.
\newblock {\em arXiv preprint arXiv:2411.15738}, 2024.

\bibitem{yu2023mm}
Weihao Yu, Zhengyuan Yang, Linjie Li, Jianfeng Wang, Kevin Lin, Zicheng Liu, Xinchao Wang, and Lijuan Wang.
\newblock Mm-vet: Evaluating large multimodal models for integrated capabilities.
\newblock {\em arXiv preprint arXiv:2308.02490}, 2023.

\bibitem{opens2v}
Shenghai Yuan, Xianyi He, Yufan Deng, Yang Ye, Jinfa Huang, Bin Lin, Chongyang Ma, Jiebo Luo, and Li~Yuan.
\newblock Opens2v-nexus: A detailed benchmark and million-scale dataset for subject-to-video generation.
\newblock {\em arXiv preprint arXiv:2505.20292}, 2025.

\bibitem{consisid}
Shenghai Yuan, Jinfa Huang, Xianyi He, Yunyuan Ge, Yujun Shi, Liuhan Chen, Jiebo Luo, and Li~Yuan.
\newblock Identity-preserving text-to-video generation by frequency decomposition.
\newblock {\em arXiv preprint arXiv:2411.17440}, 2024.

\bibitem{yue2024mmmu}
Xiang Yue, Yuansheng Ni, Kai Zhang, Tianyu Zheng, Ruoqi Liu, Ge~Zhang, Samuel Stevens, Dongfu Jiang, Weiming Ren, Yuxuan Sun, et~al.
\newblock Mmmu: A massive multi-discipline multimodal understanding and reasoning benchmark for expert agi.
\newblock In {\em Proceedings of the IEEE/CVF Conference on Computer Vision and Pattern Recognition}, pages 9556--9567, 2024.

\bibitem{zhai2023sigmoid}
Xiaohua Zhai, Basil Mustafa, Alexander Kolesnikov, and Lucas Beyer.
\newblock Sigmoid loss for language image pre-training.
\newblock In {\em Proceedings of the IEEE/CVF international conference on computer vision}, pages 11975--11986, 2023.

\bibitem{zhang2025unified}
Jihai Zhang, Tianle Li, Linjie Li, Zhengyuan Yang, and Yu~Cheng.
\newblock Are unified vision-language models necessary: Generalization across understanding and generation.
\newblock {\em arXiv preprint arXiv:2505.23043}, 2025.

\bibitem{MagicBrush}
Kai Zhang, Lingbo Mo, Wenhu Chen, Huan Sun, and Yu~Su.
\newblock Magicbrush: A manually annotated dataset for instruction-guided image editing.
\newblock {\em Advances in Neural Information Processing Systems}, 36:31428--31449, 2023.

\bibitem{LLaVA-NeXT}
Y~Zhang, B~Li, H~Liu, Y~Lee, L~Gui, D~Fu, J~Feng, Z~Liu, and C~Li.
\newblock Llava-next: A strong zero-shot video understanding model.
\newblock 2024.

\bibitem{zhang2025context}
Zechuan Zhang, Ji~Xie, Yu~Lu, Zongxin Yang, and Yi~Yang.
\newblock In-context edit: Enabling instructional image editing with in-context generation in large scale diffusion transformer.
\newblock {\em arXiv preprint arXiv:2504.20690}, 2025.

\bibitem{UltraEdit}
Haozhe Zhao, Xiaojian~Shawn Ma, Liang Chen, Shuzheng Si, Rujie Wu, Kaikai An, Peiyu Yu, Minjia Zhang, Qing Li, and Baobao Chang.
\newblock Ultraedit: Instruction-based fine-grained image editing at scale.
\newblock {\em Advances in Neural Information Processing Systems}, 37:3058--3093, 2024.

\bibitem{zhu2023languagebind}
Bin Zhu, Bin Lin, Munan Ning, Yang Yan, Jiaxi Cui, HongFa Wang, Yatian Pang, Wenhao Jiang, Junwu Zhang, Zongwei Li, et~al.
\newblock Languagebind: Extending video-language pretraining to n-modality by language-based semantic alignment.
\newblock {\em arXiv preprint arXiv:2310.01852}, 2023.

\end{thebibliography}

\end{document}